\documentclass[sigconf]{acmart}

\AtBeginDocument{%
  }

\usepackage{booktabs}
\usepackage{multirow}

\setcopyright{acmlicensed}
\copyrightyear{2018}
\acmYear{2018}
\acmDOI{XXXXXXX.XXXXXXX}

\acmConference[Conference acronym 'XX]{Make sure to enter the correct
  conference title from your rights confirmation email}{June 03--05,
  2018}{Woodstock, NY}

\acmISBN{978-1-4503-XXXX-X/2018/06}

\begin{document}

\title{Are We Winning the Wrong Game?
Revisiting Evaluation Practices for Long-Term Time Series Forecasting}

\author{Thanapol Phungtua-eng}

\affiliation{%
  \institution{Rajamangala University of Technology Tawan-ok
}
  \country{Thailand}}
\email{thanapol\_ph@rmutto.ac.th}

\author{Yoshitaka Yamamoto}
\affiliation{%
  \institution{Shizuoka University}
  \city{Shizuoka}
  \country{Japan}}
\email{yyamamoto@inf.shizuoka.ac.jp}

\renewcommand{\shortauthors}{T. Phungtua-eng and Y. Yamamoto}

\begin{abstract}
Long-term time series forecasting (LTSF) is widely recognized as a central challenge in data mining and machine learning. LTSF has increasingly evolved into a benchmark-driven “GAME,” where models are ranked, compared, and declared state-of-the-art based primarily on marginal reductions in aggregated pointwise error metrics such as MSE and MAE. Across a small set of canonical datasets and fixed forecasting horizons, progress is communicated through leaderboard-style tables in which lower numerical scores define success. In this GAME, what is measured becomes what is optimized, and incremental error reduction becomes the dominant currency of advancement. We argue that this metric-centric regime is not merely incomplete, but structurally misaligned with the broader objectives of forecasting. In real-world settings, forecasting often prioritizes preserving temporal structure—trend stability, seasonal coherence, robustness to regime shifts, and supporting downstream decision processes. Optimizing aggregate pointwise error does not necessarily imply modeling these structural properties. As a result, leaderboard improvement may increasingly reflect specialization in benchmark configurations rather than a deeper understanding of temporal dynamics. This paper revisits LTSF evaluation as a foundational question in data science: what does it mean to measure forecasting progress? We propose a multi-dimensional evaluation perspective that integrates statistical fidelity, structural coherence, and decision-level relevance. By challenging the current metric monoculture, we aim to redirect attention from winning benchmark tables toward advancing meaningful, context-aware forecasting.

\end{abstract}



\received{20 February 2007}
\received[revised]{12 March 2009}
\received[accepted]{5 June 2009}

\maketitle

\section{When Benchmark Success Becomes the Objective of Long-Term Time Series Forecasting}
Long-term time series forecasting (LTSF) is a central challenge in major data mining and machine learning venues \cite{kim2025surveytsf}. Driven by applications such as energy forecasting, traffic monitoring, and climate analysis, LTSF research has focused on improving predictive performance over long horizons. This momentum has led to the rapid development of increasingly complex forecasting models and large-scale benchmark-based evaluation practices.

A major turning point in LTSF research was the introduction of Informer \cite{haoyietal-informer-2021}. Informer not only proposed a new architecture but also standardized evaluation protocols on widely adopted benchmarks such as Electricity Transformer (ETT) and Weather. These protocols evaluate models on fixed benchmark datasets under predefined forecasting horizons (e.g., 96, 192, 336, and 720 time steps) using pointwise error metrics including mean squared error (MSE) and mean absolute error (MAE), aggregated over forecasting windows. By adopting aggregated MSE and MAE as the primary evaluation metrics, Informer-style evaluation established a reproducible but narrowly defined notion of forecasting success \cite{qiu2024tfb,qiao2026itstimegenerationtime}.

Over time, this paradigm has solidified into what we describe as the \textbf{“GAME”} of LTSF: a benchmark-centered evaluation regime in which models compete under shared experimental settings and success is determined by relative numerical improvement. Performance tables increasingly operate as implicit leaderboards, where marginal reductions in aggregated MSE and MAE translate directly into claims of superiority. In this GAME, what is measured does not merely evaluate progress—it \textit{defines} it. Architectural design, training strategies, and empirical narratives progressively align with leaderboard optimization rather than critical examination of the structural validity of forecasts. Table~\ref{tab:table1} exemplifies this reporting pattern, which continues to dominate evaluation practices at major venues, including KDD, AAAI, and NeurIPS.

\begin{table}[t]
\centering
\caption{Illustrative benchmark-style reporting for LTSF across multiple datasets. Pointwise error metrics (MSE/MAE) are reported for multiple horizons (96, 192, 336, 720) across three recent methods.}
\label{tab:table1}
\resizebox{\linewidth}{!}{
\begin{tabular}{ll|cc|cc|cc|c}
\toprule
\multirow{2}{*}{Dataset} & \multirow{2}{*}{Pred.} &
\multicolumn{2}{c|}{\begin{tabular}[c]{@{}c@{}}Method A\\(2025)\end{tabular}} &
\multicolumn{2}{c|}{\begin{tabular}[c]{@{}c@{}}Method B\\(2025)\end{tabular}} &
\multicolumn{2}{c|}{\begin{tabular}[c]{@{}c@{}}Method C\\(2025)\end{tabular}} &
\multirow{2}{*}{\begin{tabular}[c]{@{}c@{}}Older methods\\ (earlier years)\end{tabular}} \\
& & MSE & MAE & MSE & MAE & MSE & MAE & \\
\midrule
\multirow{4}{*}{ETTm1}
& 96  & 0.311 & 0.354 & 0.312 & 0.349 & 0.312 & 0.351 & $\cdots$ \\
& 192 & 0.352 & 0.379 & 0.355 & 0.372 & 0.361 & 0.378 & $\cdots$ \\
& 336 & 0.381 & 0.401 & 0.392 & 0.395 & 0.392 & 0.401 & $\cdots$ \\
& 720 & 0.439 & 0.439 & 0.466 & 0.431 & 0.453 & 0.438 & $\cdots$ \\
\midrule
\multirow{4}{*}{ETTm2}
& 96  & 0.170 & 0.254 & 0.165 & 0.250 & 0.173 & 0.258 & $\cdots$ \\
& 192 & 0.236 & 0.298 & 0.230 & 0.293 & 0.238 & 0.301 & $\cdots$ \\
& 336 & 0.294 & 0.336 & 0.292 & 0.333 & 0.297 & 0.338 & $\cdots$ \\
& 720 & 0.388 & 0.394 & 0.381 & 0.384 & 0.393 & 0.394 & $\cdots$ \\
\midrule
\multicolumn{9}{c}{$\vdots$} \\
\bottomrule
\end{tabular}
}
\end{table}

Standardization has improved comparability. Yet it has also concentrated attention on a small set of metrics and datasets, producing a form of \textit{metric monoculture}. In this setting, incremental reductions in averaged MSE and MAE increasingly shape research incentives, potentially overshadowing structural properties of forecasts such as trend preservation, seasonal coherence, robustness to regime shifts, and decision-level impact. As a result, benchmark optimization risks becoming the implicit objective of the forecasting rather than a means of assessing forecasting quality.

\subsection{A Big and Bold Question: Is the Evaluation Game Aligned with Forecasting Objectives?}

Many LTSF studies present evaluation tables similar to Table~\ref{tab:table1}, and subsequent work often follows comparable experimental protocols and benchmark comparisons. As a result, such tables have become a dominant reference point for assessing model performance. This raises fundamental questions: 

\textit{Are we optimizing for meaningful long-term forecasting, or for leaderboard superiority? What exactly are we optimizing? Who defines what counts as “progress”? When a model reduces MSE by 0.003, what has truly improved?}

To illustrate this concern, consider a hypothetical scenario in which methods developed three or four years from now consistently achieve lower MSE and MAE than those reported in Table~\ref{tab:table1}, thereby “winning” the benchmark competition. Would such numerical improvements truly signal better long-term forecasting, or merely deeper adaptation to the rules of this evaluation game? 

If models are rewarded for outperforming prior baselines under fixed experimental conditions, what incentives does this create? Does leaderboard ascent reflect broader generalization, or increasing specialization to canonical datasets?

This distinction is critical. The question is not whether models can achieve lower error, but whether such improvements necessarily correspond to deeper forecasting insight.

\subsection{When Evaluation Shapes Objectives}

Within the LTSF GAME, benchmark tables increasingly act as unofficial leaderboards. Numerical rankings, even when separated by marginal differences, influence how models are perceived, cited, and adopted. Do research narratives increasingly emphasize incremental gains because those gains determine comparative ranking?

If performance gains primarily arise from adapting to benchmark statistics rather than from a deeper understanding of temporal dynamics, the field risks \textit{mistaking leaderboard improvement for genuine forecasting advancement}. Note that, we use the term temporal dynamics to refer to the structured evolution of a time series over time, including patterns such as trend, seasonality, regime transitions, and other systematic changes in behavior.

At a broader level, we must ask: what kind of scientific culture does the current evaluation GAME incentivize? Are we cultivating a deeper theoretical understanding of temporal dynamics, or refining strategies to outperform a fixed benchmark configuration? If the community collectively optimizes for leaderboard superiority, does benchmark victory become an implicit proxy for scientific truth? And if so, is that sufficient for a field devoted to understanding temporal dynamics in time series?

Unlike prior discussions that focus primarily on alternative error metrics or expanded benchmark collections, this paper reframes LTSF evaluation as an incentive structure that shapes research objectives. Benchmark-driven competition can implicitly redefine forecasting success by rewarding marginal improvements in aggregated pointwise errors. In response, we advocate a broader evaluation perspective that considers multiple dimensions of forecasting quality, including statistical fidelity, structural coherence, and decision-level relevance. Our goal is not to replace benchmarks, but to encourage evaluation practices that better reflect the broader objectives of time series analysis.

\section{Why This Question Matters Now} \label{sec:sec2}

In recent years, several questions have surfaced within the LTSF community. Prior work has challenged the assumed superiority of complex Transformer-based architectures, demonstrating that simpler linear models can achieve competitive or even superior performance under standard evaluation protocols \cite{Zeng2023dlinear,das2023longterm}. Other studies have questioned whether widely adopted benchmark protocols, such as standardized dataset splits, fixed forecasting horizons, and uniform error aggregation schemes—adequately reflect real-world forecasting challenges \cite{qiu2024tfb,qiao2026itstimegenerationtime,bergmeir2024fundamental}.

These discussions suggest growing re-examination of LTSF assumptions, yet most current work still operates within the same benchmark-driven evaluation regime.

The \textit{urgency of this question} arises from the scale and momentum of current research. As models become more complex and benchmark protocols more entrenched, incremental metric improvements risk solidifying a narrow definition of progress. If left unexamined, this direction may reinforce optimization toward benchmark-specific gains rather than toward a deeper understanding of temporal dynamics and real-world forecasting objectives.

In this section, we examine three factors that motivate rethinking evaluation within the LTSF \textbf{GAME}: benchmark-constrained model development, differing structural interpretations of forecasting, and incentive effects driven by metric dominance. Together, these observations highlight two underexplored aspects of forecasting quality—structural coherence and decision-level relevance—that are often overlooked in benchmark-centered evaluation.

\subsection{Benchmark-Driven Model Development}
The benchmark datasets used in the LTSF \textbf{GAME} form a relatively small and standardized evaluation environment. While such benchmarks enable reproducible comparison across methods, they represent only a limited subset of the diverse temporal behaviors encountered in real-world forecasting applications.

This concern is consistent with observations by Brigato et al. \cite{brigato2026there}, who argue that newly proposed models may increasingly rely on complex architecture and dataset-specific experimental setups and evaluation protocols, thereby reinforcing benchmark-centered optimization. Similarly, Qiu et al.~\cite{qiu2024tfb} demonstrate that time series exhibit substantial heterogeneity, including varying degrees of trend strength, seasonality, regime shifts, and distributional change.

Taken together, these observations suggest a structural risk: when evaluation concentrates on a small, fixed benchmark set, apparent progress may increasingly reflect specialization to benchmark characteristics rather than improved robustness across diverse forecasting scenarios.

\subsection{One Dataset, Multiple Interpretations}
Even when evaluated on the same dataset, forecasting depends on assumptions about how future values evolve. In classical time series analysis, seasonal--trend decomposition represents a series as a combination of trend, seasonal, and residual components \cite{cleveland1990stl}. Other frameworks introduce different structural views; for example, Prophet models time series using trend, seasonal, and holiday effects \cite{taylor2018forecasting}, while alternative approaches distinguish between local and global trend dynamics \cite{SMYL2025111,sophaken2026lgtdlocalglobaltrenddecomposition}. These choices reflect different interpretations of the underlying temporal dynamics.

Such differences become particularly important when the time series exhibits abrupt changes. Some models allow sharp trend shifts, while others assume the trend evolves smoothly and treat abrupt variations as anomalies or irregular disturbances \cite{wen2019robuststl}. If a forecasting model intentionally smooths abrupt changes under the assumption that they are anomalous, its predictions may deviate from observed values. However, under certain analytical objectives, this structurally consistent forecast may be preferable to one that closely follows abrupt fluctuations.

Within the GAME, evaluation is largely reduced to numerical proximity between predictions and observed values. As a result, forecasts based on different structural assumptions are judged primarily by pointwise error, even when they imply distinct views of the underlying temporal dynamics. This reveals a fundamental limitation of the prevailing evaluation objective: the issue is not simply that forecasting models rely on different structural assumptions, but that pointwise metrics cannot determine which of these differences is most relevant to the analytical goal at hand.

\subsection{Incentive Effects and Metric Dominance (Forecasting $\neq$ Curve Fitting)} 
Real-world forecasting tasks are not always concerned with reproducing every fluctuation in observed data. In many applications, decision-making depends on capturing broader structural behavior—such as trend direction, seasonal consistency, or regime shifts—rather than matching high-frequency noise. For example, in energy planning, traffic management, or financial analysis, stakeholders may prioritize stable trend estimation over exact replication of short-term irregularities. In such contexts, a forecast that abstracts transient disturbances may be more informative than one that closely tracks noisy observations.

A simple yet widely adopted example is the use of exponential moving averages (e.g., EMA-20, EMA-50, and EMA-100) in financial time series analysis. These indicators are commonly employed to extract persistent trend signals and inform portfolio allocation or trading decisions, even though they are not designed to minimize pointwise forecasting error \cite{lemperiere2014centuriestrendfollowing,park2007profitability}. Empirical evidence suggests that trend-following strategies derived from such signals can generate long-term economic value when evaluated in terms of return and risk-adjusted performance, which differ from MSE or MAE.

This contrast points to a broader tension: minimizing pointwise deviation on benchmark datasets is not necessarily equivalent to producing forecasts that are most useful in practice. A model that better supports structural interpretation or downstream decision-making may perform worse under averaged error metrics, while a model that more closely matches noisy observations may appear superior in benchmark tables.

If evaluation metrics become the \textit{primary currency of progress in the community}, model development may increasingly prioritize statistical fit over structural usefulness and decision relevance. In this way, forecasting risks being reduced to curve fitting rather than serving as a tool for understanding temporal dynamics and informing action.

In the next section, we discuss how redefining evaluation objectives may help realign LTSF research with its broader analytical goals and outline frontier-level directions for future work.

\section{How Does This Paper Push the Frontier?}
Building on the motivations discussed in Section~\ref{sec:sec2}, we propose a forward-looking, a three-dimensional perspective on forecasting performance centered on statistical fidelity, structural coherence, and decision-level relevance. We argue that the frontier of LTSF lies not in another incremental metric or model variant, but in rethinking how forecasting performance is defined and assessed.

\subsection{Beyond Pointwise Error Metrics}
Recent works \cite{hewamalage2023forecast,bergmeir2024fundamental,kim2025surveytsf} highlight how evaluation design and benchmark construction can significantly influence empirical conclusions and model rankings. These observations suggest that forecasting evaluation should be understood as a multi-dimensional problem rather than a single-number optimization task. Consequently, we advocate for a structural rethinking of how forecasting performance is defined and assessed. We propose that future LTSF evaluation be organized along three complementary dimensions:

\begin{itemize}
  \item \textbf{Statistical Fidelity:} Forecasts should be evaluated for how closely they match observed values under standardized conditions. Scaled errors (e.g., MASE) and probabilistic measures (e.g., CRPS) can improve comparability across datasets and settings. Although not accuracy measures, computational metrics such as inference time and training efficiency remain relevant for practical deployment.

  \item \textbf{Structural Coherence:} Forecasts should be evaluated in terms of their alignment with underlying temporal structure, including trend preservation, seasonal consistency, and robustness to regime shifts. When models explicitly incorporate decomposition mechanisms, evaluation should examine whether these structural components are meaningfully preserved in prediction. Structural characteristics may be assessed using quantitative diagnostics such as trend strength, seasonal strength, distributional properties, and randomness tests. Qiu et al. \cite{qiu2024tfb} show how such diagnostics can systematically characterize dataset properties. Building on this perspective, we can apply those diagnostics to forecast outputs and evaluate whether the predicted series preserves meaningful temporal structure.

  \item \textbf{Decision-Level Relevance:} In many practical settings, forecasting serves as an input to downstream analysis or action. Evaluation should therefore consider whether forecast outputs are useful for domain-specific objectives, such as robustness under distributional change, asymmetric error costs, operational planning, or risk-aware decision-making. In some cases, forecasting quality may be better assessed through its impact on related tasks such as anomaly detection, imputation, or other application-specific goals.

\end{itemize}

In this view, these dimensions reframe evaluation not as a single leaderboard ranking, but as a structured assessment of forecasting behavior under diverse analytical objectives. An immediate practical question then follows: how should such evaluation be reported and interpreted in empirical studies?

\subsection{From Leaderboard Ranking to Diagnostic Reporting}
We advocate that the next step for LTSF evaluation should move beyond leaderboard ordering toward diagnostic reporting. Instead of asking which model achieves the lowest averaged error, evaluation should examine under what structural conditions a model succeeds or fails. 

Encouragingly, recent benchmarking initiatives such as TSFM-Bench \cite{Li2025tsfmbench}, TFB \cite{qiu2024tfb}, and TIME \cite{qiao2026itstimegenerationtime} demonstrate growing awareness of dataset auditing, metric diversity, and structured evaluation settings, including few-shot and zero-shot scenarios. Building on this momentum, we advocate evaluation protocols that emphasize interpretability, regime-specific performance, and structural robustness rather than aggregate ranking alone.

Another promising direction is window-level analysis, where forecasted results are evaluated against ground truth at the level of individual sliding windows rather than being aggregated into global average metrics. This approach enables explicit examination of performance variability across regimes, structural conditions, and noise levels.

For example, certain windows may contain anomalies, abrupt shifts, or extreme outliers, leading to larger deviations between forecasts and ground truth. In such cases, elevated error values may reflect structural irregularity rather than model inadequacy.

Conversely, if a window contains only a few isolated outliers, a structurally consistent forecasting model may deliberately smooth these irregular points instead of matching them exactly. Under pointwise error metrics, such behavior would increase MSE or MAE, even though the forecast may better represent the underlying temporal dynamics.

\begin{figure*}[t]
  \centering
  \includegraphics[width=\linewidth]{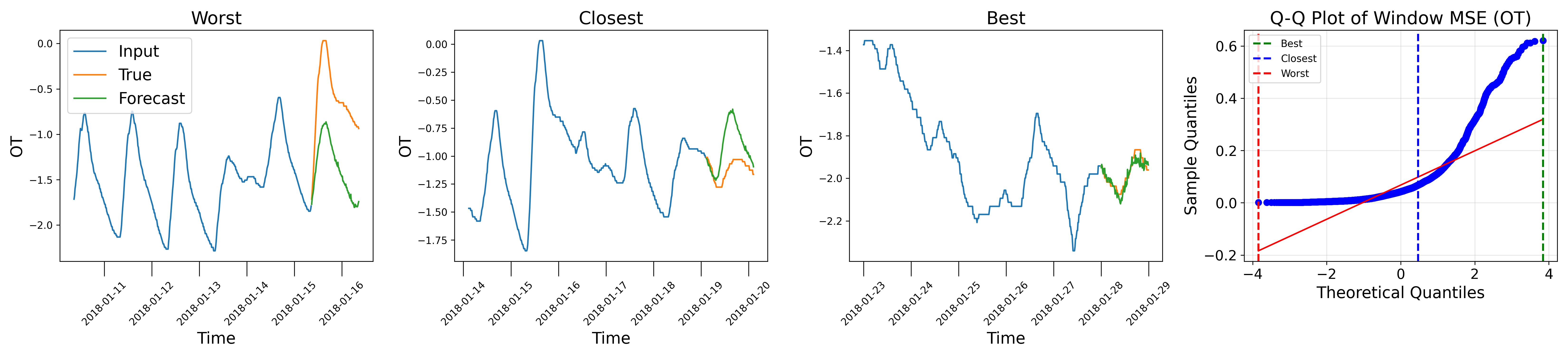}
  \caption{Forecasting results for the oil temperature (OT) series in the ETTm2 dataset using the DLinear \cite{Zeng2023dlinear} model. The figure shows three representative sliding windows: the highest MSE (Worst), a window closest to the mean MSE (Closest), and the lowest MSE (Best). For each window, recent input observations before the cutoff are plotted together with the ground truth and forecasted values over the prediction horizon. The figure highlights the variability of forecasting performance across windows that may be hidden by aggregated error metrics.}
  \label{fig:figuer1}
\end{figure*}

Treating all windows uniformly under aggregated metrics risks obscuring heterogeneity and misinterpreting model behavior. To better visualize this variability, window-level error distributions can be examined using diagnostic tools such as Q–Q plots. Figure~\ref{fig:figuer1} illustrates how windows containing structural irregularities deviate significantly from the dominant error distribution.

Diagnostic reporting is not intended to replace aggregate metrics, but to contextualize them. By shifting from averaged evaluation to distribution-aware diagnostics, forecasting assessment becomes a structural analysis problem rather than a single-score comparison. 

In this view, progress in LTSF is defined not merely by lower averages, but by a principled understanding of when, where, and why models succeed or fail across diverse structural conditions. This perspective also challenges the common assumption that a single forecasting model can universally dominate across datasets and application scenarios.

\subsection{No Universal Champion: Domain-Specific and Context-Dependent Performance}
Within the benchmark-driven evaluation paradigm popularized by Informer \cite{haoyietal-informer-2021}, models are often presented as “state-of-the-art” based on marginal improvements in averaged MSE and MAE across shared benchmark datasets.

However, recent analyses suggest that such dominance is rarely universal. Brigato et al.~\cite{brigato2026there} report that no single forecasting model consistently outperforms others across all datasets, prediction horizons, and structural conditions. These findings challenge the implicit assumption that a universally superior LTSF model exists.

From this perspective, forecasting models should be evaluated relative to clearly defined domain objectives rather than solely on their ability to achieve marginal improvements in aggregated benchmark metrics. Models are typically designed with specific inductive biases and assumptions tailored to particular application contexts.

For example, an LTSF model is deployed to monitor electricity loads in an operational setting. It may prioritize stability, robustness to regime shifts, and interpretability over minimal MSE on the ETTh dataset. Similarly, a model may be deployed for exchange rate forecasting. In such a setting, the quality of its forecasts may be judged by their contribution to risk-aware decision-making, rather than solely on pointwise error reduction.

This perspective suggests that evaluation should shift from absolute leaderboard ranking toward conditional performance analysis under domain-specific criteria. Rather than searching for a universal champion, the field may benefit from identifying which models perform reliably under specific structural conditions and decision objectives.

\section{What Would Success Look Like?}
If the perspective advanced in this paper is adopted, success would no longer be defined solely by incremental reductions in pointwise error metrics. Instead, progress in LTSF would be measured by the degree to which the objectives of evaluation align with real-world forecasting goals.

In such a future, leaderboard competition would give way to structural understanding. Models would be assessed not only by numerical proximity to observed values, but by their ability to preserve meaningful temporal dynamics—such as trend, seasonality, and regime transitions—and by their robustness across domain-specific contexts.

Evaluation would become context-aware and objective-driven. Forecast quality would be judged relative to its analytical usefulness, decision relevance, and interpretability within specific application domains, rather than by aggregate ranking alone.

Ultimately, success would be reflected in a cultural shift within the LTSF community: contributions would be valued for advancing understanding of temporal dynamics and generating practical insight, not merely for achieving marginal benchmark gains. The impact of this idea would therefore lie not in introducing another metric, but in reshaping how meaningful progress in forecasting research is defined and recognized.

In this sense, forecasting progress should not be measured only by how closely models match past observations, but by how effectively they help us understand and act upon temporal dynamics.

\balance
\bibliographystyle{ACM-Reference-Format}
\bibliography{BSKDD2026}

\clearpage
\appendix
\section{Suplementary Materials}
\subsection{Implementation Details}
We used the NeuralForecast library\footnote{https://nixtlaverse.nixtla.io/} in Python to facilitate reproducibility. All experiments were run on an NVIDIA RTX 5060 Ti GPU with 16\,GB VRAM and an AMD EPYC 7352 24-core CPU.

\subsection{Window-Level Diagnostics}
\begin{figure*}[t]
  \centering
  \includegraphics[width=\linewidth]{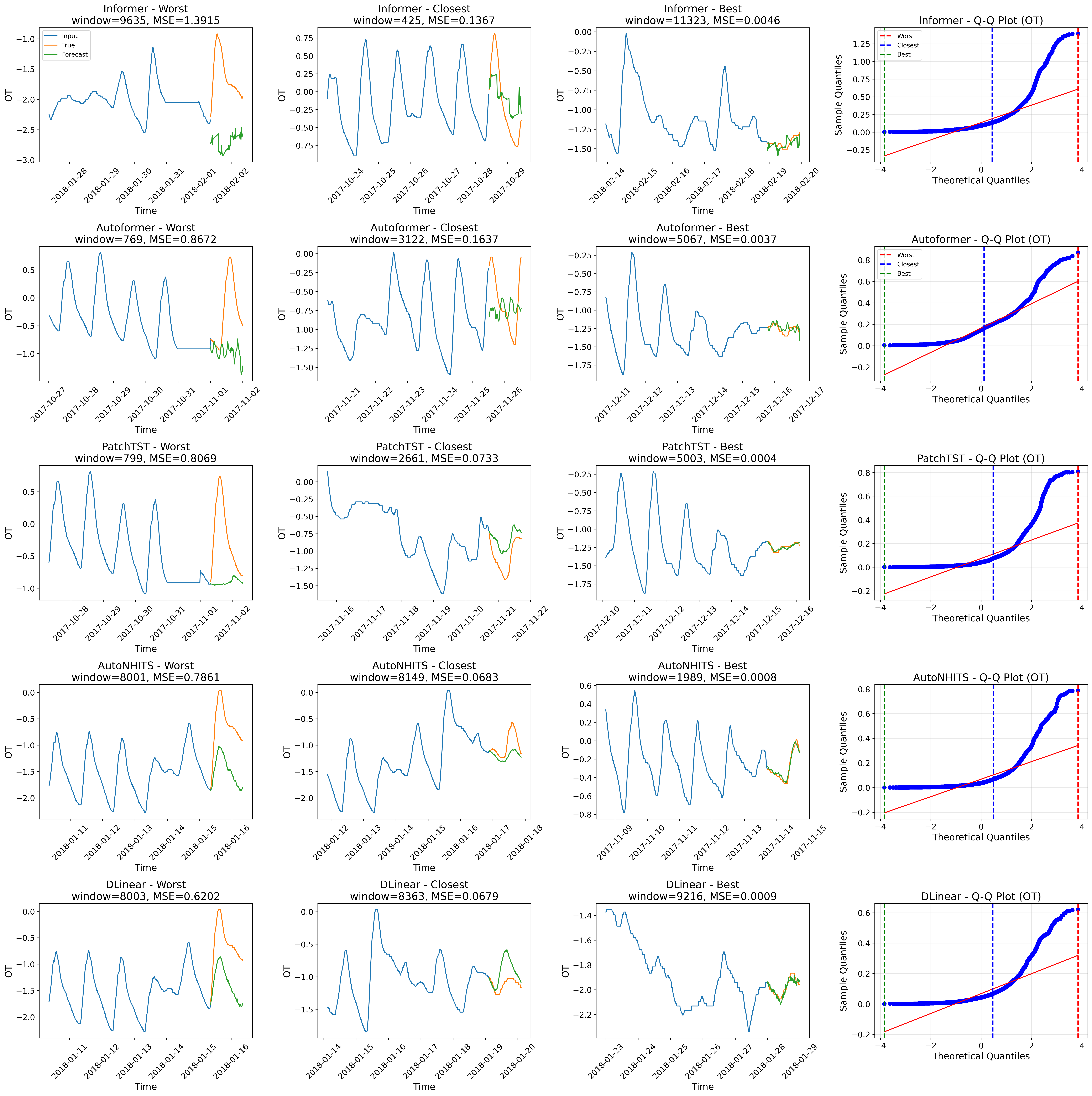}
  \caption{Forecasting results for the \textcolor{red}{oil temperature (OT)} series in the ETTm2 dataset across models. The figure shows three representative sliding windows: the highest MSE (Worst), a window closest to the mean MSE (Closest), and the lowest MSE (Best). For each window, recent input observations before the cutoff are plotted together with the ground truth and forecasted values over the prediction horizon. The figure highlights the variability of forecasting performance across windows that may be hidden by aggregated error metrics.}
  \label{fig:figuer2}
\end{figure*}

\begin{figure*}[t]
  \centering
  \includegraphics[width=\linewidth]{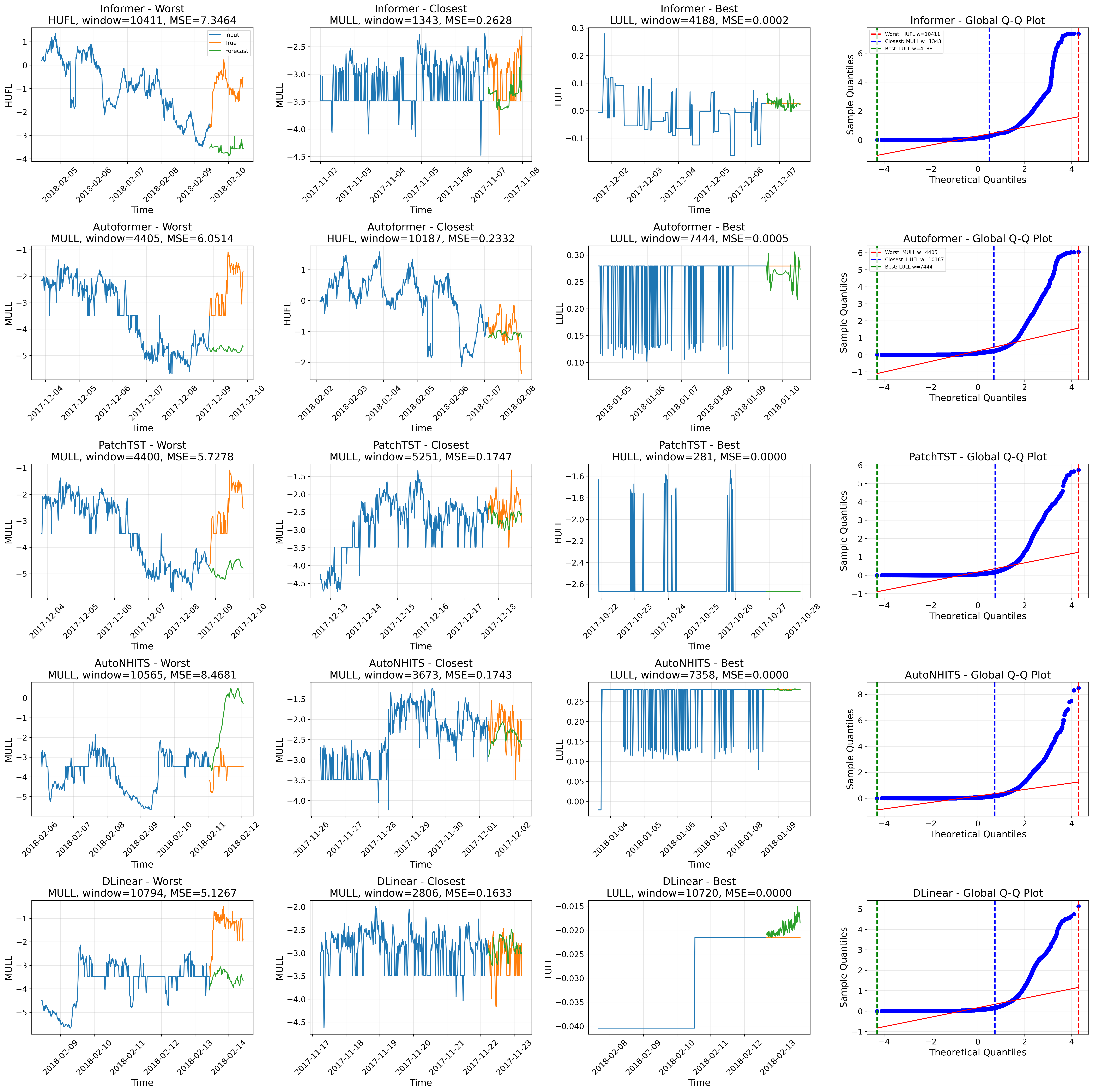}
  \caption{Forecasting results for \textcolor{red}{all variables} in the ETTm2 dataset across models.The figure shows three representative sliding windows: the highest MSE (Worst), a window closest to the mean MSE (Closest), and the lowest MSE (Best). For each window, recent input observations before the cutoff are plotted together with the ground truth and forecasted values over the prediction horizon. The figure highlights the variability of forecasting performance across windows that may be hidden by aggregated error metrics.}
  \label{fig:figuer3}
\end{figure*}

Following the results shown in Figure~\ref{fig:figuer1}, we provide additional window-level examples on the ETTm2 dataset using OT and other variables. These comparisons include several state-of-the-art methods, namely Informer \cite{haoyietal-informer-2021}, Autoformer \cite{wu2021autoformer}, DLinear \cite{Zeng2023dlinear}, NHITS \cite{challu2023nhits}, and PatchTST \cite{Yuqietal-2023-PatchTST}. The corresponding results are shown in Figures~\ref{fig:figuer2}--\ref{fig:figuer3}.

\end{document}